\documentclass[conference]{IEEEtran}
\IEEEoverridecommandlockouts
\usepackage{amsmath,amssymb,amsfonts}
\usepackage{algorithmic}
\usepackage{algorithm}
\usepackage{graphicx}
\usepackage{subcaption}
\usepackage{textcomp}
\usepackage{xcolor}
\usepackage{multirow}
\usepackage{graphicx}

\usepackage{tikz}
\usepackage[backend=bibtex,%
    sorting=nty,%
    giveninits=true,
    eprint=false,%
    url=false,doi=false,isbn=false]{biblatex}
\usepackage{common}

\makeatletter
\def\ps@IEEEtitlepagestyle{
  \def\@oddfoot{\mycopyrightnotice}
  \def\@oddhead{\hbox{}\@IEEEheaderstyle\leftmark\hfil\thepage}\relax
  \def\@evenhead{\@IEEEheaderstyle\thepage\hfil\leftmark\hbox{}}\relax
  \def\@evenfoot{}
}
\def\mycopyrightnotice{
  \begin{minipage}{\textwidth}
  \centering \scriptsize
  Copyright~\copyright~2022 IEEE. Personal use of this material is permitted. Permission from IEEE must be obtained for all other uses, in any current or future media, including\\reprinting/republishing this material for advertising or promotional purposes, creating new collective works, for resale or redistribution to servers or lists, or reuse of any copyrighted component of this work in other works by sending a request to pubs-permissions@ieee.org.
  \end{minipage}
}
\makeatother

\def\BibTeX{{\rm B\kern-.05em{\sc i\kern-.025em b}\kern-.08em
    T\kern-.1667em\lower.7ex\hbox{E}\kern-.125emX}}
\begin{document}

\title{Sample Condensation in Online Continual Learning
\thanks{This research was supported by TEACHING, a project funded by the EU
Horizon 2020 research and innovation programme under GA n. 871385}
}
\author{\IEEEauthorblockN{Anonymous Authors}}

\author{\IEEEauthorblockN{Mattia Sangermano}
\IEEEauthorblockA{\textit{Computer Science Dept.} \\
\textit{University of Pisa}\\
Pisa, Italy \\
mattiasangermano1997@gmail.com}
\and
\IEEEauthorblockN{Antonio Carta}
\IEEEauthorblockA{\textit{Computer Science Dept.} \\
\textit{University of Pisa}\\
Pisa, Italy \\
antonio.carta@unipi.it}
\and
\IEEEauthorblockN{Andrea Cossu}
\IEEEauthorblockA{
\textit{Scuola Normale Superiore}\\
Pisa, Italy \\
andrea.cossu@sns.it}
\and
\IEEEauthorblockN{Davide Bacciu}
\IEEEauthorblockA{\textit{Computer Science Dept.} \\
\textit{University of Pisa}\\
Pisa, Italy \\
davide.bacciu@unipi.it}

}

\maketitle
\begin{abstract}
Online Continual learning is a challenging learning scenario where the model must learn from a non-stationary stream of data where each sample is seen only once. The main challenge is to incrementally learn while avoiding catastrophic forgetting, namely the problem of forgetting previously acquired knowledge while learning from new data. A popular solution in these scenario is to use a small memory to retain old data and rehearse them over time. Unfortunately, due to the limited memory size, the quality of the memory will deteriorate over time. In this paper we propose OLCGM, a novel replay-based continual learning strategy that uses knowledge condensation techniques to continuously compress the memory and achieve a better use of its limited size. The sample condensation step compresses old samples, instead of removing them like other replay strategies. As a result, the experiments show that, whenever the memory budget is limited compared to the complexity of the data, OLCGM improves the final accuracy compared to state-of-the-art replay strategies.
\end{abstract}

\begin{IEEEkeywords}
continual learning; online learning; replay; knowledge condensation.
\end{IEEEkeywords}

\section{Introduction}
% need for CL
Deep learning models have achieved state-of-the-art results in fields such as computer vision \cite{resnet, yolo, Imagenet} and natural language processing \cite{gpt3, bert}. However, all these results assume the presence of a static training dataset, representative of the entire data distribution. In practice, most learning environments are non-stationary, since data arrives as a stream and the underlying data distribution may change over time \cite{catastrophic-forgetting}. In such environments, the ability to learn incrementally over time, i.e., Continual Learning (CL) \cite{LML,CL_robotics,CL-RL,LrL}, is necessary. Unfortunately, most models fail to retain the knowledge about past data when they are trained on new data, a problem known as catastrophic forgetting \cite{cat-forg2,catastrophic-forgetting}.

% ocl scenario
In the literature, Online Continual Learning (OCL) is considered one of the most challenging continual learning scenarios. In OCL, input data arrives in small mini-batches (usually $1$ to $10$ samples at a time) and the model is trained in a single pass, unaware of any changes (boundaries) in the underlying data distribution \cite{Exstream}.
% replay-based ocl
Most strategies fail in this extreme setting, except for replay-based strategies \cite{MIR, Gdumb, icarll}, which keep a fixed-size memory (or buffer) of previous samples to use for rehearsal. During training, old and new samples are interleaved to learn incrementally while mitigating catastrophic forgetting. 
% naive policies
Replay-based strategies use very simple operations to manipulate their memory: a policy to select which samples to add and remove, and how to sample from the memory during training. Most strategies use simple policies such as random selection or greedy policies \cite{RR, MIR, Gdumb}.
% need for adaptive policies
Unfortunately, naive usage of the bounded rehearsal memory is very inefficient, and replay-based strategies are forced to remove many elements whenever the memory is updated. Over time, the memory updates will remove useful information about the previous tasks. Since the replay buffer is a component of the model, we believe that the model should learn how to manipulate its buffer and design adaptive policies, optimized for the specific learning environment.
\begin{figure}
    \centering
    \includegraphics[width=0.48\textwidth]{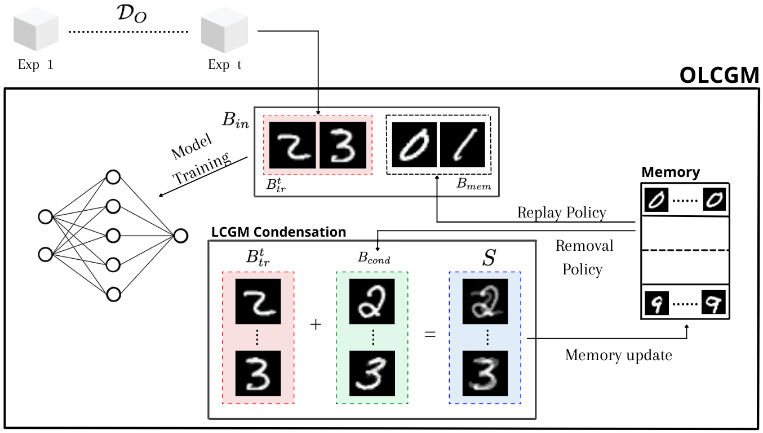}
    \caption{High-level illustration of OLCGM. For each set of samples received in input OLCGM replay an additional set of images from the memory. Together with the incoming sample, the replayed images are used to update the parameters of the classification model. Then, to enhance the information in the memory, the received examples are condensed together with a set of examples from the memory and saved within it.}
    \vspace{-1.5em}
  \label{fig:OLCGM}
\end{figure}

% paper proposal
In this paper, we address the aforementioned desiderata by proposing a novel replay-based strategy for OCL, called Online Linear Condensation with Gradient Matching (OLCGM)\footnote{Code available at: https://github.com/MattiaSangermano/OLCGM}. A high level overview of the strategy is shown in Figure \ref{fig:OLCGM}. OLCGM updates the replay buffer with a novel operation: the \emph{sample condensation}. Instead of removing samples, at each memory update \emph{a subset of the examples are condensed together in a novel synthetic sample}. This approach allows to mitigate the forgetting caused by the removal of old samples since the synthetic sample will compress information about multiple samples into a single one.
% paper results
The experimental results in OCL scenarios show that, when the memory size is small compared to the variability in the original data, naive removal policies suffer from catastrophic forgetting since they need to discard samples from the memory very frequently to make space for new ones.

\section{Related Works}
The main objective of continual learning is to learn incrementally over time without catastrophic forgetting \cite{cat-forg2, catastrophic-forgetting}. Most continual learning strategies in the literature can be grouped into three different families. The first family includes regularization strategies \cite{EWC,LwF,SI}, namely strategies that use penalty terms to impose constraints on the weights update. In this way, the CL algorithm is guided towards solutions that generalize over all the tasks addressed, avoiding more specialized solutions that would result in catastrophic forgetting. Secondly, architectural-based strategies \cite{ExpertGate, PackNet, PNN, HAT} change the model's architecture, either statically or dynamically, as new tasks arise. These may include the addition of new units or layers, as well as freezing selected parameters to prevent changes. Finally, replay-based strategies \cite{MIR,GEM, Gdumb,BiC} are CL algorithms that store samples or information about the past tasks in an external memory. The memory is used during the training of new tasks in order to rehearse the old knowledge and therefore to avoid forgetting.
While in general there are many tradeoffs that make some of these solutions more preferable to others, it is unquestionable that in most Online Continual Learning scenarios the only strategies that consistently achieve reliable results are replay-based strategies \cite{MIR,Gdumb}. For example, \cite{Gdumb} keeps the memory balanced between the classes such that every time data from a new class arrives it is also added into the memory and an equal number of patterns from the other classes is removed. The selection of examples to add and remove from the memory is done using a greedy policy. The model used for the classification is trained using only the examples in memory. Notice that while this strategy achieves a good performance, it is questionable whether it can be considered an online strategy since it requires retraining the model from scratch before the evaluation. Nonetheless, it provides a simple baseline to compare against. Instead, the objective of Maximally Interfered Retrieval (MIR) \cite{MIR} is to augment each incoming input batch with the examples that most interfere with the input samples available from the memory. Even though these methods are able to achieve promising results, their use of greedy policies to remove and add examples from the memory is suboptimal. In general, in an OCL scenario, we expect that over time the memory will lose information about the previous tasks. Indeed, the main difference between these strategies and our work is the way in which examples are removed from the memory. In fact, the strategy we propose aims to reduce the forgetting of the knowledge present in the examples that need to be removed when new ones are added. In our case, this is done through the use of a new condensation algorithm that, in addition to compressing the information inside the memory, adaptively selects and removes all examples that contain useless information. A similar work has been proposed in \cite{CCM,DCGM}, but differently from them, in our case the condensation algorithm is optimised to work in an OCL scenario. The condensation algorithms in \cite{CCM,DCGM} do not take into account the OCL constraints, therefore they are not applicable to such setting.

\section{Sample Condensation}
In an OCL scenario, a CL algorithm processes a continuous and non-stationary stream of experiences $\mathcal{D}_{O}$, where each \emph{experience} is a small set of samples, typically at most $10$ samples. At each iteration $t$ a new set of examples $B_{tr}^t = \{ (x,y)^i \}_{i=1}^{L} \sim p_{t}$ becomes available, where $x$ and $y$ are the input images and class labels respectively, while $p_{t}$ is the distribution from which $B_{tr}^t$ is drawn and $L$ is the number of samples received at each iteration. The distribution $p_{t}$, may be equal to the previous distribution $p_{t-1}$ or it may be a different one. Indeed, the algorithm is unaware of any changes in the data distribution. In addition, $L$ may change over time. For these reasons, the algorithm must be able to continuously adapt itself. These properties make the OCL scenario much more difficult to solve than other CL scenarios since the task boundaries are not defined, unlike multi-task \cite{CL_robotics, mtl} or class-incremental \cite{three-scen, cil, cil2} scenarios. Finally, the combination of a long stream and very small batches of data makes most of the algorithms designed for CL to fail in OCL, either because of their increased computational cost or because of catastrophic forgetting.
\subsection{Replay in Online Continual Learning}
Previous works applied to the OCL scenario mainly focus on the design of replay-based strategies, namely algorithms that use an external memory to store and rehearse past data in order to increase the robustness of the model to the catastrophic forgetting problem. The external memory often stores raw examples, but other alternatives are possible as well \cite{A-GEM,Exstream, GEM, icarll, ExAI}.  The high-level pseudocode for a basic replay strategy in an OCL scenario is described by Algorithm \ref{alg:OCL-replay} below:
\begin{algorithm}[H]
    \centering
    \caption{Replay strategy in OCL scenario}
    \label{alg:OCL-replay}
    \begin{algorithmic}[1]
        \REQUIRE data stream $\mathcal{D}_O$, classifier $f$, memory $\mathcal{M}$
        \FOR{$ B_{tr}^t \in  \mathcal{D}_O$}
            \STATE Sample $B_{mem} \sim \mathcal{M}$ \label{line:memory-sampling}
            \STATE $B_{in} \leftarrow B_{tr}^t \cup B_{mem}$ \label{line:Bin}
            \STATE Update $\theta \leftarrow \texttt{train}(f,B_{in})$ \label{line:model-adapt}
            \STATE Sample $B_{in} \sim B_{tr}^t$ \label{line:int-sampling}
            \STATE $\mathcal{M}.\texttt{update}(B_{in})$ \label{line:mem-upd}
        \ENDFOR
    \end{algorithmic}
\end{algorithm}
Lines \ref{line:memory-sampling}, \ref{line:int-sampling} and \ref{line:mem-upd} describe the interactions of the training algorithm with the external memory $\mathcal{M}$: sampling from the memory, choosing which examples to add, and updating the buffer. The exact implementation of the replay in Algorithm \ref{alg:OCL-replay} depends on the definition of these three operations. $\mathcal{M}$ can be partitioned into separate buckets (e.g. by class label) and filled using both raw examples or different kind of information useful to the model during the train phase. Each of the lines \ref{line:memory-sampling}, \ref{line:int-sampling}, \ref{line:mem-upd} is characterized by a different policy: line \ref{line:memory-sampling} uses a sampling policy to decide which examples to retrieve from the memory in order to supplement the mini-batch received as input, line \ref{line:int-sampling} choose the examples to be inserted in memory through the use of an insertion policy and line \ref{line:mem-upd} through a removal policy chooses which examples to remove from memory in order to free space for the new ones.  The current literature focuses on the design of optimal heuristics for the three aforementioned policies \cite{RR,MIR}. However, all of this heuristics ignore a fundamental problem: \emph{how can we avoid losing information each time we update the external memory?} Because of computational constraints, the external memory will have a limited size. As a result, whenever we update the memory with new examples, we also need to decide which samples to keep and which to remove. Over time, the online update of the external memory will lead to a gradual forgetting of critical information included in the examples removed from the memory. This limitation will always happen in an online scenario as soon as the complexity and variability of the past data in the stream will become larger than what the memory can store. Intuitively, if the algorithm runs for a long enough time, it will always have to remove informative examples at some point. In this section, we design an algorithm to mitigate this problem. The main idea behind the algorithm is the design of a \emph{sample condensation algorithm that allows to compress multiple examples into a single one}, avoiding the need for removing data from the buffer. While previous attempts exist for class-incremental learning scenarios \cite{CCM,DCGM}, all of them are unfeasible for OCL. In particular, the large number of parameters of the synthetic images, makes them too computationally expensive. In the following, we will introduce Dataset Condensation and will show how to perform the sample condensation for replay strategies in the OCL scenario.

\subsection{Dataset Condensation}
\label{sec:DC}
The objective of the Dataset Condensation \cite{DD} is to compress the knowledge of an entire dataset into few synthetic training images by optimizing the synthetic images for fast adaptation. After the condensation, the synthetic data can be used for training in place of the original data. Since the synthetic dataset is much smaller than the original dataset, we can reduce the size of the training data (or an external memory) with a small drop in accuracy. Following \cite{DCGM}, let $B = \{(x_i,y_i)\}_{i=1}^{|B|}$ be a dataset, where $x_i $ and $y_i$ are the input data and the label of the $i$-th entry of the dataset. The goal of the condensation is to generate a set of synthetic samples with their corresponding labels, $S = \{(s_i,y_i)\}_{i=1}^{|S|}$ where $|S| \ll |B|$, such that the synthetic samples can be used to train a neural network $f$ to achieve the same performance as if the same network was trained on the original dataset $B$. This objective can be expressed as
\begin{equation}
    \mathbb{E}_{(x,y) \sim P_{B}} [\ell(f_{\theta^{B}}(x),y)] \ \simeq \  \mathbb{E}_{(x,y) \sim P_{B}} [\ell(f_{\theta^{S}}(x),y)],
    \label{eq:gen-perf}
\end{equation}
where $P_B$ is the data distribution, $\ell(\cdot,\cdot)$ is a task specific loss, $f$ is the classification model, $\theta^{B}$ and $\theta^{S}$ are the network parameters obtained training the network on the datasets $B$ and $S$ respectively. The point at which condensation comes into play in Equation \ref{eq:gen-perf} is the way in which the synthetic images S are optimised. The optimisation of the set $S$ and therefore $\theta^S$  would bring the right-hand side closer to the left-hand side of Equation \ref{eq:gen-perf}. Depending on the objective to be pursued, different condensation algorithms may be produced. In \cite{DD} the authors propose to learn the syntethic images $S$ such that the model $f_{\theta^S}$ trained on the set S minimizes the training loss over the original dataset $D$, namely:
\begin{align}
    S^*= & \operatornamewithlimits{argmin}\limits_{S} \mathcal{L}^{B}(\theta^{S}) 
    \label{eq:opt-DD1} 
    \\
    \text{subject to} \quad  \theta^{S} & = \ \operatornamewithlimits{argmin}\limits_{\theta} \mathcal{L}^{S}(\theta),
    \label{eq:opt-DD2} 
    \\
    \mathcal{L}^{S}(\theta)= \frac{1}{|S|} & \sum \limits_{(x,y) \in S} \ell\big(f_{\theta}(x),y\big)
    \label{eq:opt-DD3} 
\end{align}
The optimal synthetic images $S^*$ are produced by alternating the inner and outer meta-optimisation steps illustrated by Equations \ref{eq:opt-DD2} and \ref{eq:opt-DD1} respectively. The inner step has the objective to find the optimal weights $\theta^S$ induced by the synthetic images $S$. The weights $\theta^S$ are then used during the outer loop to quantify the quality of the synthetic images produced so far. The synthetic set $S$ during the outer-optimization step is optimised in such a way that the weights $\theta^S$ obtained from $S$ during the next iteration of the inner loop increase the performance on the original dataset $D$. Despite the goodness of this approach in producing high quality synthetic sets, the multiple optimization step of $\theta^S$ in Equation \ref{eq:opt-DD2} leads to unroll the recursive computation graph in order to recover the gradients required for the tuning of $S$ in Equation \ref{eq:opt-DD1}. The unrolling is computationally expensive and therefore does not scale to large models. In order to alleviate this limitation, \cite{DCGM} proposes the Dataset Condensation with Gradient Matching algorithm (DCGM) that learns the synthetic images by minimizing the distance in parameter space of $\theta^B$ and $\theta^S$. Therefore, in addition to achieving a performance similar to $\theta^B$, their approach also steers the optimization of $\theta^S$ towards $\theta^B$. This objective can be achieved by solving the following optimization problem:
\begin{equation}
    \label{eq:opt-LCGM}
    \begin{aligned}
        \operatornamewithlimits{min}\limits_{\mathcal{S}} \mathbb{E} \Bigg[ \sum \limits_{t=1}^{T} \mathcal{D} &\big( \nabla_{\theta^{S}}\mathcal{L}^{S}(\theta_t^{S}), \nabla_{\theta^{S}}\mathcal{L}^{B}(\theta_t^{S}\big) \Bigg],\\
        \mathcal{D}(A,B) =& \sum \limits_{i}^{L} \Big(1 - \frac{A_i\cdot B_i}{\lVert A_i \rVert \cdot \lVert B_i \rVert}\Big)
    \end{aligned}
\end{equation}
Where $A_i$ and $B_i$ are the flattened vectors of the gradients corresponding to the $i$-th output nodes. The idea behind Equation \ref{eq:opt-LCGM} is to force the learning trajectory in the parameter space of $\theta^S$ to be as similar as possible to that of $\theta^B$. Equation \ref{eq:opt-LCGM} has the advantage over Equation \ref{eq:opt-DD1} of not requiring the unrolling of the computational graph for the optimization of $S$ during each iteration $t$. This means that DCGM is fast and memory efficient and therefore is able to scale also with large models. The reader is referred to \cite{DCGM} for a detailed analysis of the steps taken to achieve Equation \ref{eq:opt-LCGM} from Equation \ref{eq:opt-DD1}.

\subsection{Linear Combination with Gradient Matching (LCGM)}
In this section, we introduce and describe the Linear Combination with Gradient Matching (LCGM), a novel technique which makes the external memory condensation more efficient and scalable to Online Continual Learning scenarios. While DCGM drastically reduces the cost of the dataset condensation compared to a naive Dataset Distillation \cite{DD}, it remains too expensive to be applied directly in an OCL scenario. The number of parameters to optimize during the condensation is the total number of pixels of the synthetic images, which results in a slow convergence rate for the condensation step.
In order to make the condensation step tractable in OCL, we propose to create synthetic images that are defined as a linear combination of the input images $S=\mathcal{W}B$, where $\mathcal{W} \in \mathbb{R}^{n \times m}$ are the coefficients of the linear combination. The coefficients $\mathcal{W}$ are learnt through the use of gradient matching (Section \ref{sec:DC}). Therefore, the coefficients are derived by:
\begin{equation}
    \label{eq:def}
    \begin{aligned}
        \operatornamewithlimits{min}\limits_{\mathcal{W}} \mathbb{E} \Bigg[ \sum \limits_{i=1}^{T} \mathcal{D} \big( \nabla_{\theta^{S}}\mathcal{L}^{S}&(\theta_i^{S}), \nabla_{\theta^{S}}\mathcal{L}^{B}(\theta_i^{S}) \big) \Bigg] \\
        \text{subject to} \quad  \theta^{S} & = \ \operatornamewithlimits{argmin}\limits_{\theta} \mathcal{L}^{S}(\theta),
        \\
        \mathcal{L}^{S}(\theta)= \frac{1}{|S|} & \sum \limits_{(x,y) \in S} \ell\big(f_{\theta}(x),y\big).
    \end{aligned}
\end{equation}
Algorithm \ref{alg:LCGM} shows the pseudocode for the external memory condensation performed by LCGM. After randomly initialising the coefficients $\mathcal{W}$, the synthetic image set $S$ is generated by a linear combination of the input images $B$. The coefficients are filtered through a mask $M$, whose objective is to define from which input images each synthetic image must be composed of. Then, after taking a mini-batch of synthetic images and one of input images, the network $f$ is used to compute the loss over both the input samples ($\mathcal{L}^B$), the synthetic samples ($\mathcal{L}^S$) and their gradients w.r.t. $\theta$. The gradients $\nabla_{\theta_t}\mathcal{L}_c^S(\theta_t)$, $\nabla_{\theta_t}\mathcal{L}_c^B(\theta_t)$ are used to compute the gradient matching loss which is employed to update the coefficients $\mathcal{W}$ at the end of each outer loop.  After generating the new set of condensed images, the parameters $\theta$ are updated by minimizing the loss $\mathcal{L}^S$ with learning rate $\eta_{\theta}$ for $I$ steps.

\subsubsection{Masking}
\begin{figure}
    \centering
        \begin{subfigure}{0.15\textwidth}
            \includegraphics[width=1\textwidth]{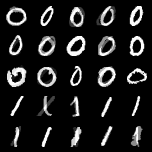}
            \subcaption{Using matrix mask.}
        \end{subfigure}
        \begin{subfigure}{0.15\textwidth}
            \includegraphics[width=1\textwidth]{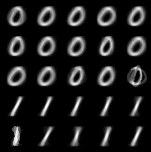}
            \subcaption{Without matrix mask.}
        \end{subfigure}
        \caption{Results of condensed MNIST images with and without the use of the matrix mask}
        \label{fig:mask}
        \vspace{-1.5em}
\end{figure}

The matrix $M$ is a bit-wise mask used to constrain the relationship between input and synthetic images: $S = (\mathcal{W} \odot M) \cdot B$. So if the $i$-th row of $\mathcal{W}$ defines the coefficients used to generate image $i$, then all the input images $j$ such that $M_{ij} = 0$ will not be taken into account for the generation of the synthetic image $i$. Masking allows to control how the condensation process transfers information from the input images into the synthetic ones and allows to craft custom policies. For example, without any masking all the synthetic images are used to optimize the input ones. While this may seem optimal, this approach may reduce the diversity of the synthetic images. Since all the generated images use the same input images they end up very similar to each other, which means the buffer will be full of redundant images. A result of this choice is shown in Figure \ref{fig:mask}, which clearly shows how the images without masking tend to be optimized into a single mode.

Instead, the use of $M$ introduces the possibility of crafting custom policies that limit the choice of which images to condense together. For example, we may want to generate each synthetic image by selecting a subset of the input images that are very similar to each other. In the subsequent experiments, we use the mask matrix to ensure that each synthetic image is composed of two images: one real and one synthetic (generated from the previous iterations).

\subsubsection{Clipping and normalization}
During the preliminary experiments, it was noticed that when the coefficients get negative or high values the resulting images reduce the learning capacity of OLCGM making it to underfit. For this reason whenever the condensed image set $S$ has to be generated, the coefficients $\mathcal{W}$ are adjusted via clipping and normalization operations. Clipping forces all coefficients to be greater or equal than 0:
\begin{equation}
    \mathcal{W}_{i,j} = \texttt{max}(\mathcal{W}_{i,j},0).
    \label{eq:clipping}
\end{equation}
Given that the synthetic images are generated through a linear combination of the input images, constraint \ref{eq:clipping} is needed since coefficients with a negative value would cause the associated images to have reverted colors and would results in the generation of meaningless images. The values of the linear combination coefficients of OLCGM and the pixel values resulting from a condensation algorithm that optimises images pixel by pixel have a different meaning. In the case of the linear combination, the coefficient can be interpreted as a representation of the amount of information contained in an image. For this reason, it is worthwhile to have a procedure able to filter out all images that according to the condensation algorithm contain useless or redundant information, i.e. those having a coefficient equal to 0. Therefore, clipping can be seen as a learnable mechanism to remove images from the buffer.

After clipping, the coefficients are normalised to prevent them from becoming too large. Without normalisation, the coefficients could increase dramatically during the optimization process, and this would lead to the generation of images that make the training of the classifier very unstable. We avoided to use regularization techniques because they would not guarantee the generation of images with pixels on the same scale as the real ones. Images with different scales of values have a different impact on the model update. Since we want to avoid that some images are considered a priori more influential than others, we preferred to normalize the coefficients in such a way that the sum of the coefficients of each synthetic image is equal to $1$ i.e: $\mathcal{W}_{i,j} = \frac{\mathcal{W}_{i,j}}{\sum_{l} \mathcal{W}_{i,l}}.$
\subsubsection{Number of parameters}
If we consider that \cite{DCGM}, in order to generate the synthetic set, updates the images pixel by pixel, then the number of parameters to train is equal to $n \cdot l$, where $n$ is the number of synthetic images and $l$ the number of pixels of each figure. In LCGM, on the other hand, the number of parameters to be trained using gradient matching are those present in the $\mathcal{W}$ matrix, namely $n \cdot m$ coefficients where $m$ is the number of images to condense. Therefore LCGM has $\frac{l}{m}$ times fewer parameters than DCGM, i.e. the ratio between the size of an image and the number of images to be condensed. Considering that in the smallest computer vision datasets, such as MNIST, each image has $28 \times 28$ pixels, and considering that in an Online Continual Learning scenario the number of input samples available at a certain timestep is small (usually no more than 10 samples), then the number of parameters of LCGM is at nearly about two orders of magnitude less than the ones in DCGM.
\begin{algorithm}
    \caption{LCGM memory condensation}
    \label{alg:LCGM}
    \begin{algorithmic}[1]
    \REQUIRE B: Set of images to be condensed, network parameters initialization $\theta$, the number of classes $C$, classifier $f$, number of outer loops $T$, number of inner loops $I$, learning rates for updating weights $\eta_{\theta}$ and matrix coefficients $\eta_{\mathcal{W}}$, mask $M$
    \STATE $\theta_0 \leftarrow \theta$ \label{line:param-init}
    \STATE $S \leftarrow (\mathcal{W} \odot M) \cdot B$ \label{line:mask1}
    \FOR{$t = 0,...,T - 1$ }
        \FOR{$c = 0,...,C - 1$ }
            \STATE Sample mini-batches $\mathcal{B}_c^B \sim B$ and $\mathcal{B}_c^S \sim S$
            \STATE $\mathcal{L}_c^B \leftarrow \frac{1}{|\mathcal{B}_c^B|}\sum_{(x,y) \in \mathcal{B}_c^B} \ell \big(f_{\theta_t}(x),y\big)$ 
            \STATE $\mathcal{L}_c^S \leftarrow \frac{1}{|\mathcal{B}_c^S|}\sum_{(x,y) \in \mathcal{B}_c^S} \ell \big(f_{\theta_t}(x),y\big)$ 
            \STATE $\mathcal{L}_{distill}^c \leftarrow D\big(\nabla_{\theta_t}\mathcal{L}_c^S(\theta_t),\nabla_{\theta_t}\mathcal{L}_c^B(\theta_t)\big)$ 
        \ENDFOR
        \STATE $\mathcal{W} \leftarrow \mathcal{W} - \eta_S \cdot \nabla_{\mathcal{W}} \sum_i \mathcal{L}_{distill}^i$
        \STATE $S \leftarrow (\mathcal{W} \odot M) \cdot B$ \label{line:mask2} 
        \STATE $\theta_{t+1} \leftarrow \texttt{opt-alg}_{\theta}(\mathcal{L}^S(\theta_t),I,\eta_{\theta})$ \label{line:13}
    \ENDFOR
    \RETURN $S$
    \end{algorithmic}
\end{algorithm}
\setlength{\textfloatsep}{0pt}
\vspace{-0.5em}
\subsection{LCGM in Online Continual Learning (OLCGM)}
\begin{algorithm}
    \centering
    \caption{LCGM in Online Continual Learning scenario}
    \label{alg:LCGM-OCL}
    \begin{algorithmic}[1]
        \REQUIRE train stream $\mathcal{D}_O$, classifier $f$,  memory $\mathcal{M}=\{  \mathcal{M}_c\}_{c=0}^C$
        \FOR{$ B_{tr}^i \in  \mathcal{D}_O$}
            \STATE Sample $B_{mem} \leftarrow \texttt{balanced-replay}(\mathcal{M})$
            \STATE $B_{in} \leftarrow B_{tr}, B_{mem}$
            \STATE Update $\theta \leftarrow \text{train}(f,B_{in})$
            \IF{$i \mod K$}
                \STATE Update $\mathcal{M} \leftarrow \texttt{memory-condensation}(\mathcal{M},B_{tr},i)$
            \ENDIF
        \ENDFOR
    \end{algorithmic}
\end{algorithm}

LCGM was designed with the objective of applying it within the Online Continual Learning scenario. For this reason, its computational cost has been greatly reduced compared to the cost of DCGM. Since the proposed strategy falls under replay methods, an external memory $\mathcal{M}$ is kept to store what is learned during the life cycle of the algorithm. The memory, as in \cite{Exstream, Gdumb,icarll}, is partitioned according to the target labels of the examples inside the memory. Algorithm \ref{alg:LCGM-OCL} shows how LCGM condensation is integrated in an Online Continual Learning scenario. For each set of input samples $B_{tr}$, a mini-batch of data is taken from the memory $B_{mem}$. The samples from the external memory are chosen to balance the classes of the images taken from the memory. $B_{tr}$ and $B_{mem}$ are then used to update the weights of the model $f$. At this point, with a frequency $K$, the input mini-batch $B_{tr}$ is condensed in the external memory $\mathcal{M}$ using the LCGM condensation technique described by Algorithm \ref{alg:LCGM}. In order to condense the $B_{tr}$ images, it is necessary to randomly select the same number of images from the memory so that each new image can be paired with one coming from the memory having the same class. The coupling of the images is guaranteed by the mask matrix $M$ which must fulfill the following additional constraints:
\begin{equation}
        \begin{aligned}
            \forall i,j \in \{0, \dots,|M|\}, \quad M_{i,j} \in \{0,1\}& \quad \wedge \\
            \forall i \in \{0, \dots,|M|\}, \quad \sum\limits_{j=0}^{|M|} M_{i,j} = 2 &  \quad \wedge \\
            \forall j \in \{0, \dots,|M|\}, \quad \sum\limits_{i=0}^{|M|} M_{i,j} = 1 & .
        \end{aligned}
\end{equation}
The first constraint ensures that the matrix is made up of only bits, the second that each synthetic image is composed of a linear combination of only two images and the third ensures that the images to be condensed are used to generate only one synthetic image. 

The condensation rate $K$ balances the computational cost of the whole OCL training phase and the degradation of the images in memory. Indeed, condensation after condensation the algorithm might reach the maximum limit of information that can be stored in a single image. Once this limit has been reached, condensation would only lead to image degradation and therefore would increase the forgetting of the previous learned knowledge. 

The memory during the whole training phase contains only condensed examples. This is useful because each condensed example contains more information than a real sample and therefore allows to optimise memory usage. Moreover, because of the way condensation works, the condensed examples will have a different distribution than the real images. Having a memory composed of only condensed examples ensures that the mini-batches $B_{in}$ used for the training of the model $f$ are balanced with respect to both the distribution of the data and that of the memory.

During the training phase, when an image with a new class is presented, the memory is resized in order to balance it among classes. The resize is performed by downsizing the space reserved for each class (using the LCGM condensation algorithm) and by adding an equal amount of space for the new class.

\section{Experiments}
In this section, we evaluate OLCGM in OCL scenarios and compare it against different replay-based baselines for increasing memory sizes. The development of OLCGM and the experimental phase has been implemented using the library Avalanche \cite{avalanche}.

\subsection{Benchmarks}
The effectiveness of OLCGM has been validated on the benchmarks SplitMNIST, SplitFashionMNIST and SplitCIFAR10 adapted for the Online Learning scenario. In these benchmarks the training phase is divided in experiences where within each experience only one partition of the original dataset is used to train the continual learning algorithm. The partitioning of the initial dataset is done in such a way as to respect the constraints of a Class-Incremental scenario \cite{three-scen}. The number of experiences used during all the experimental phase is equal to 5 and the task ids are not provided. In order to simulate an Online Continual Learning scenario each experience is in turn divided in a stream of smaller experiences each of size 10. Therefore, at each iteration of the training phase a new small set of images is received as input. The performance of the compared strategies are evaluated using the average accuracy (ACC) and average forgetting (AF) metrics defined as $\text{ACC} = \frac{1}{N} \sum\limits_{i=1}^N A_i^N$, $\text{AF} = \frac{1}{N-1} \sum\limits_{i=1}^{N - 1} A_i^{i} - A_i^{N - 1} ,$ where $A_i^j$ denotes the accuracy of the model on experience $i$ computed right after being trained on experience $j$.
\subsection{State-of-the-art strategies}
We compare OLCGM against three replay-based state-of-the-art strategies from the literature. The first is the random replay (RR) \cite{RR}, the simplest replay-based strategy. This baseline chooses randomly both the examples to be added and removed from the memory and the samples replayed from the memory that are used during the training phase. The second algorithm is MIR, a strategy that builds the mini-batches used for the training by retrieving from the memory the samples which are most interfered with the ones received in input. The last one is the same strategy as OLCGM with the only difference that the condensation algorithm is the one presented in \cite{DCGM} and described in Section \ref{sec:DC}. In the following, we will refer to this strategy as ODCGM.
\subsection{Model Architectures}
During the experiments conducted on the SplitMNIST and SplitFashionMNIST benchmarks we used a Multilayer Perceptron with an hidden layer of 400 neurons and RELU as activation function, while in the SplitCIFAR10 benchmark we used a smaller version of Resnet-18 as in \cite{A-GEM, GEM}. In all benchmarks, Resnet-18 and MLP classifiers are optimised by using SGD with a learning rate equal to $0.1$.
\subsection{Model Selection}
In order to choose the best hyper-parameters of each algorithm, a grid search has been made, using as validation set a subset of the training data of each dataset. In all three benchmarks, the validation set consists of 500 examples for each experience. In the case of MIR, the hyperparameters are the same as those chosen in \cite{MIR}. For ODCGM and OLCGM the only hyperparameters fine-tuned for the model selection are those present in the condensation module: the learning rate used for the generation of the synthetic images, the number of outer and inner loops. Random Replay is the only strategy that has no hyperparameters apart from those included in the internal model.
\begin{table}
   \centering
   \resizebox{0.8\linewidth}{!}{
    \begin{tabular}{|c |c |c c c| c c c|}
        \hline
          \multirow{2}*{$\lvert M \rvert$}& \multirow{2}*{Benchmark} & \multicolumn{3}{c|}{OLCGM} & \multicolumn{3}{c|}{ODCGM} \\ 
         & & Ol & Il & lr & Ol & Il & lr \\ \hline
        
        \multirow{3}*{10} & SMNIST & 100 & 1 & 0.1 & 50 & 1 & 0.1 \\
        & SFMNIST & 200 & 1 & 0.01 & 200 & 5 & 0.1 \\
        & SCIFAR10 & 100 & 1 & 0.01 & 50 & 1 & 0.1 \\ \hline
        
        \multirow{3}*{20} & SMNIST & 100 & 1 & 0.1 & 50 & 1 & 0.1 \\
        & SFMNIST & 200 & 1 & 0.01 & 50 & 5 & 0.1 \\
        & SCIFAR10 & 200 & 1 & 0.01 & 100 & 1 & 0.1 \\ \hline
        
        \multirow{3}*{50} & SMNIST & 100 & 1 & 0.1 & 50 & 5 & 0.1 \\
        & SFMNIST & 200 & 1 & 0.01 & 200 & 1 & 0.1 \\
        & SCIFAR10 & 100 & 1 & 0.01 & 50 & 1 & 0.1 \\ \hline
        
        \multirow{3}*{100} & SMNIST & 50 & 1 & 0.1 & 200 & 1 & 0.1 \\
        & SFMNIST & 200 & 1 & 0.01 & 50 & 1 & 0.1 \\
        & SCIFAR10 & 100 & 1 & 0.01 & 200 & 1 & 0.1 \\ \hline
        
        \multirow{3}*{200} & SMNIST & 50 & 1 & 0.1 & 100 & 5 & 0.1 \\
        & SFMNIST & 200 & 1 & 0.01 & 200 & 5 & 0.1 \\
        & SCIFAR10 & 100 & 5 & 0.01 & 200 & 1 & 0.1 \\ \hline
    \end{tabular}}
    \caption{Chosen hyperparameters for the experimental phase of OLCGM and ODCGM. Note that the word Split is abbreviated with the letter S in the benchmark column. The columns $Ol$, $Il$ denotes the number of iterations in the outer and inner loops, while $lr$ denotes the learning rate used for the optimization of the synthetic images.}
    \vspace{1.5em}
    \label{tab:hyperparameters}
\end{table}
Tables \ref{tab:hyperparameters} shows the best hyperparameters found for OLCGM and ODCGM, $\rvert M \lvert$ indicates the memory capacity. The condensation rate $K$ is set to 10 for SplitMNIST, SplitFashionMNIST and 50 for SplitCIFAR10. In addition, during the experiments the coefficient matrix $\mathcal{W}$ used by OLCGM for the generation of the images is randomly initialized. Initialising randomly each pixel of the ODCGM synthetic images would be unfeasible for an OCL scenario as the number of iterations required to make the synthetic images representative enough of the condensed images would be prohibitive for this scenario. Therefore, each synthetic image of ODCGM is initialized by picking randomly one of the image to be condensed.

\subsection{Results}
\begin{table*}
    \begin{minipage}{0.5\linewidth}
        \resizebox{\textwidth}{!}{
        \begin{tabular}{c c c c c c c }
        $\lvert$ \textbf{M} $\rvert$ & & 10 & 20 & 50 & 100 & 200 \\ \hline \hline
        
        \multirow{2}*{\textbf{ORR}} & ACC $\uparrow$ & $28.6 \pm 2.1$ & $37.8 \pm 3.0$ & $56.1 \pm 3.4$ & $69.4 \pm 2.1$ & $80.0 \pm 1.9$  \\
        & AF $\downarrow$ & $88.4 \pm 2.6$ & $77.3 \pm 3.7$ & $54.6 \pm 4.1$ & $37.8 \pm 2.6$ & $24.5 \pm 2.5$  \\ \hline
        
        \multirow{2}*{\textbf{MIR}} & ACC $\uparrow$ & $29.5 \pm 1.0$ & $41.0 \pm 1.6$ & $59.9 \pm 2.1$ & $\textbf{71.8} \pm \textbf{1.4}$ & $\textbf{81.7} \pm \textbf{0.8}$  \\
        & AF $\downarrow$ & $87.1 \pm 1.2$ & $72.5 \pm 2.0$ & $48.9 \pm 2.7$ & $33.9 \pm 1.7$ & $21.3 \pm 1.0$  \\ \hline
        
        \multirow{2}*{\textbf{OLCGM}} & ACC $\uparrow$ & $\textbf{37.3} \pm \textbf{2.9}$ & $\textbf{48.2} \pm \textbf{3.0}$ & $\textbf{63.4} \pm \textbf{2.6}$ & $\textbf{71.8} \pm \textbf{2.1}$ & $78.2 \pm 2.9$  \\
        & AF $\downarrow$ & $78.0 \pm 3.5$ & $64.4 \pm 3.7$ & $45.4 \pm 3.2$ & $34.7 \pm 2.6$ & $26.5 \pm 3.6$  \\ \hline
        
        \multirow{2}*{\textbf{ODCGM}} & ACC $\uparrow$ & $31.8 \pm 2.5$ & $39.2 \pm 3.6$ & $56.8 \pm 3.4$ & $69.7 \pm 2.2$ & $78.7 \pm 2.7$  \\
        & AF $\downarrow$ & $84.6 \pm 3.1$ & $75.6 \pm 4.5$ & $53.7 \pm 4.2$ & $37.6 \pm 2.7$ & $26.1 \pm 3.5$\\ \hline
        
        \end{tabular}
        }
        \caption{Average accuracy and forgetting on SplitMNIST benchmark, averaged over 15 runs (in bold are highlighted the values having the highest ACC metric with respect to the memory capacity).}
        \label{tab:SOTA-SMNIST}
    \end{minipage}
    \hfill
    \quad
    \begin{minipage}{0.5\linewidth}
        \resizebox{\textwidth}{!}{
        \begin{tabular}{c c c c c c c c}
        $\lvert$ \textbf{M} $\rvert$ & & 10 & 20 & 50 & 100 & 200 \\ \hline \hline
        
        \multirow{2}*{\textbf{ORR}} & ACC $\uparrow$ & $34.2 \pm 1.7$ & $40.8 \pm 3.2$ & $54.3 \pm 2.2$ & $64.4 \pm 1.4$ & $\textbf{72.1} \pm \textbf{1.0}$  \\
        & AF $\downarrow$ & $80.9 \pm 2.0$ & $72.6 \pm 4.0$ & $55.5 \pm 2.8$ & $42.4 \pm 1.8$ & $32.2 \pm 1.6$ \\ \hline
        
        \multirow{2}*{\textbf{MIR}} & ACC $\uparrow$ & $37.7 \pm 2.1$ & $43.9 \pm 1.5$ & $57.9 \pm 1.6$ & $65.3 \pm 1.7$ & $70.8 \pm 1.2$ \\
        & AF $\downarrow$ & $75.9 \pm 2.5$ & $68.0 \pm 2.1$ & $46.9 \pm 2.6$ & $33.8 \pm 2.4$ & $28.4 \pm 1.6$  \\ \hline
        
        \multirow{2}*{\textbf{OLCGM}} & ACC $\uparrow$ & $\textbf{41.9} \pm \textbf{3.4}$ & $\textbf{52.1} \pm \textbf{2.8}$ & $\textbf{62.3} \pm \textbf{1.4}$ & $\textbf{67.8}\pm \textbf{1.0}$ & $71.1 \pm 1.2$ \\
        & AF $\downarrow$ & $71.2 \pm 4.3$ & $58.2 \pm 3.5$ & $45.1 \pm 1.8$ & $37.7 \pm 1.4$ & $33.0 \pm 1.8$  \\ \hline
        
        \multirow{2}*{\textbf{ODCGM}} & ACC $\uparrow$ & $35.6 \pm 1.6$ & $41.4 \pm 2.9$ & $54.7 \pm 2.3$ & $63.6 \pm 1.9$ & $71.5 \pm 1.4$ \\
        & AF $\downarrow$ & $79.2 \pm 2.0$ & $71.8 \pm 3.6$ & $54.9 \pm 2.9$ & $43.2 \pm 2.6$ & $32.9 \pm 1.9$ \\ \hline
    \end{tabular}}
    \caption{Average accuracy and forgetting on SplitFashionMNIST benchmark, averaged over 15 runs (in bold are highlighted the values having the highest ACC metric with respect to the memory capacity). }
     \label{tab:SOTA-SFASHION}
    \end{minipage} 
\end{table*}
    
\begin{table}
    \centering
    \resizebox{\linewidth}{!}{
    \begin{tabular}{c c c c c c c}
        $\lvert$ \textbf{M} $\rvert$ & & 10 & 20 & 50 & 100 & 200 \\ \hline \hline
        
        \multirow{2}*{\textbf{ORR}} & ACC $\uparrow$ & $18.9 \pm 0.4$ & $18.9 \pm 0.6$ & $20.1 \pm 0.9$ & $22.0 \pm 1.6$ & $26.3 \pm 2.5$ \\
        & AF $\downarrow$ & $86.5 \pm 1.0$ & $86.9 \pm 1.2$ & $85.0 \pm 1.2$ & $82.0 \pm 2.8$ & $75.9 \pm 3.2$  \\ \hline
        
        \multirow{2}*{\textbf{MIR}} & ACC $\uparrow$ & $18.3 \pm 0.4$ & $18.8 \pm 0.3$ & $\textbf{20.3} \pm \textbf{1.0}$ & $22.7 \pm 1.1$ & $\textbf{28.6} \pm \textbf{1.1}$ \\
        & AF $\downarrow$ & $74.9 \pm 3.1$ & $75.4 \pm 2.6$ & $67.2 \pm 3.1$ & $59.5 \pm 2.4$ & $49.1 \pm 1.7$  \\ \hline
        
        \multirow{2}*{\textbf{OLCGM}} & ACC $\uparrow$ & $\textbf{19.0} \pm \textbf{0.5}$ & $18.9 \pm 0.7$ & $20.1 \pm 1.2$ & $22.4 \pm 1.7$ & $24.1 \pm 1.7$\\
        & AF $\downarrow$ & $85.8 \pm 1.2$ & $86.2 \pm 1.3$ & $84.4 \pm 1.9$ & $81.3 \pm 3.1$ & $78.8 \pm 3.4$  \\ \hline
        
        \multirow{2}*{\textbf{ODCGM}} & ACC $\uparrow$ & $18.8 \pm 0.5$ & $\textbf{19.0} \pm \textbf{0.7}$ & $20.0 \pm 1.2$ & $\textbf{22.8} \pm \textbf{2.1}$ & $25.8 \pm 1.6$  \\
        & AF $\downarrow$ & $86.3 \pm 1.1$ & $86.5 \pm 1.3$ & $84.4 \pm 2.3$ & $81.6 \pm 3.0$ & $76.8 \pm 2.6$  \\ \hline
        
    \end{tabular}
    }
    \caption{Average accuracy and forgetting on SplitCIFAR10 benchmark, averaged over 15 runs (in bold are highlighted the values having the highest ACC metric with respect to the memory capacity).}
    \label{tab:SOTA-SCIFAR10}
    \vspace{1.5em}
\end{table}
Tables \ref{tab:SOTA-SMNIST}, \ref{tab:SOTA-SFASHION} and \ref{tab:SOTA-SCIFAR10} show the average accuracy (ACC) and average forgetting (AF) of the four strategies. Firstly, it can be seen that despite its simplicity RR is highly competitive. In contrast to what was presented in \cite{MIR}, in our experiments RR is able to achieve similar performance to MIR. OLCGM shows improved performance with respect to MIR, especially when the available memory capacity is low compared to the complexity of the data. Indeed, in this cases OLCGM is capable to improve the average accuracy of MIR up to 26\%. The advantage of OLCGM is graduallly reduced when the memory capacity increases, thus suggesting that when there is limited memory it is critical to be able to condense the largest amount of information in few images. When the memory is large, instead, it is more important to manage appropriately its content since it already stores enough information to avoid forgetting. This is most likely the reason why MIR performs better when the memory increases.

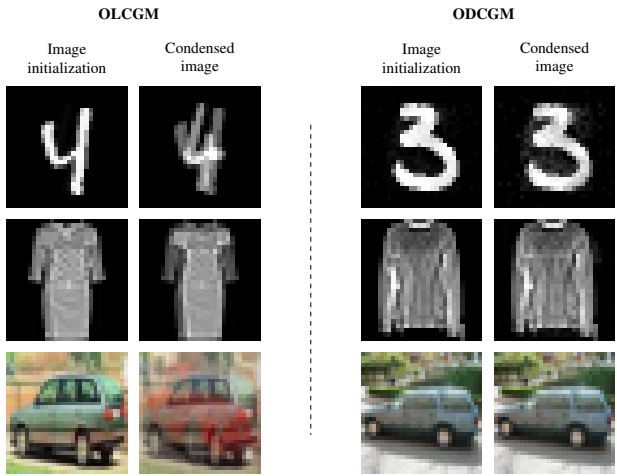
\begin{figure}[H]
    \centering
    \resizebox{0.45\textwidth}{!}{
        \begin{tikzpicture}
        \node[align=center, font=\bfseries] at (1.4,3) {OLCGM};
        \node[align=center] at (0,2) {Image\\ initialization};
        \node[align=center] at (3.0,2) {Condensed\\ image};
        % MNIST OLCGM
        \node[inner sep=0pt] (mnist_start_olcgm) at (0,0)
            {\includegraphics[width=.15\textwidth]{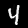}};
        \node[inner sep=0pt] (mnist_end_olcgm) at (3,0)
            {\includegraphics[width=.15\textwidth]{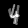}};
        % FASHION OLCGM
        \node[inner sep=0pt] (mnist_start_olcgm) at (0,-3.0)
            {\includegraphics[width=.15\textwidth]{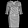}};
        \node[inner sep=0pt] (mnist_end_olcgm) at (3,-3.0)
            {\includegraphics[width=.15\textwidth]{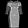}};
        % CIFAR10 OLCGM
        \node[inner sep=0pt] (mnist_start_olcgm) at (0,-6.0)
            {\includegraphics[width=.15\textwidth]{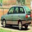}};
        \node[inner sep=0pt] (mnist_end_olcgm) at (3,-6.0)
            {\includegraphics[width=.15\textwidth]{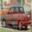}};
        
        \draw [dashed] (5.5,0.5) -- (5.5,-6.5);
        
        \node[align=center, font=\bfseries] at (9.4,3) {ODCGM};
        
        \node[align=center] at (8,2) {Image\\ initialization};
        \node[align=center] at (11.0,2) {Condensed\\ image};
        
        % MNIST OGM
        \node[inner sep=0pt] (mnist_start_olcgm) at (8,0)
            {\includegraphics[width=.15\textwidth]{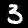}};
        \node[inner sep=0pt] (mnist_end_olcgm) at (11,0)
            {\includegraphics[width=.15\textwidth]{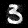}};
        
        % FASHION OGM
        
        \node[inner sep=0pt] (mnist_start_olcgm) at (8,-3.0)
            {\includegraphics[width=.15\textwidth]{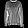}};
        \node[inner sep=0pt] (mnist_end_olcgm) at (11,-3.0)
            {\includegraphics[width=.15\textwidth]{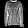}};
            
        % CIFAR10 OGM
        
        \node[inner sep=0pt] (mnist_start_olcgm) at (8,-6.0)
            {\includegraphics[width=.15\textwidth]{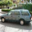}};
        \node[inner sep=0pt] (mnist_end_olcgm) at (11,-6.0)
            {\includegraphics[width=.15\textwidth]{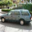}};
    \end{tikzpicture}}
    \caption{OLCGM and ODCGM synthetic images after the initialization and after their optimization}
    \label{fig:condensation-comp}
\end{figure}
The same pattern described above arises when comparing OLCGM with ODCGM. That is, with limited memory capacity, OLCGM achieves higher performance, whereas when increasing the memory size this advantage disappears. At a first glance, this would suggest that ODCGM with high memory capacities is able to produce a better representation of knowledge in memory than OLCGM. In fact, when looking at the generated images in Figure \ref{fig:condensation-comp}, we can see that the condensation method of ODCGM collapses to images that are almost the same as when they were initialised. Therefore, the condensed images are very similar to the real ones. Having images in the buffer that are very similar to the real ones makes ODCGM behave like RR. In fact their performances are very similar. Overall, we can conclude that whenever the memory size is large enough, removing random samples like done by RR is sufficient. However, whenever the data stream is much more complex than what the limited memory can represent with raw samples, we see that the sample condensation of OLCGM outperforms RR, MIR, and ODCGM.

\section{Conclusions}
In this paper, we studied the problem of updating the rehearsal memory of replay-based algorithms in Online Continual Learning scenarios. OCL is still an unsolved problem, and most strategies proposed in the literature are replay-based. The majority of these strategies rely on fixed and naive random policies to add or remove examples. However, while these simple policies may be successfull in simple toy streams, like those used in the literature, they use the memory buffer inefficiently. Instead, we proposed an alternative solution that compresses different samples together instead of removing them, thus allowing a better use of the rehearsal buffer via an adaptive compression.
The results show that, whenever the memory buffer is small, i.e. in those settings where naive policies fail, compression-based strategies like the proposed one allow to improve the average accuracy of the model and to reduce the forgetting.
We hope these results will encourage more research in adaptive methods to manage the rehearsal buffer. In the future, we plan to extend the proposed strategy to make it scale to more difficult benchmarks and further reduce its computational cost.

\printbibliography

@article{LML,
  title={Lifelong machine learning},
  author={Chen, Zhiyuan and Liu, Bing},
  journal={Synthesis Lectures on Artificial Intelligence and Machine Learning},
  volume={12},
  number={3},
  pages={1--207},
  year={2018},
  publisher={Morgan \& Claypool Publishers}
}

@article{Imagenet,
  title={Imagenet large scale visual recognition challenge},
  author={Russakovsky, Olga and Deng, Jia and Su, Hao and Krause, Jonathan and Satheesh, Sanjeev and Ma, Sean and Huang, Zhiheng and Karpathy, Andrej and Khosla, Aditya and Bernstein, Michael and others},
  journal={International journal of computer vision},
  volume={115},
  number={3},
  pages={211--252},
  year={2015},
  publisher={Springer}
}

@article{bert,
  title={Bert: Pre-training of deep bidirectional transformers for language understanding},
  author={Devlin, Jacob and Chang, Ming-Wei and Lee, Kenton and Toutanova, Kristina},
  journal={arXiv preprint arXiv:1810.04805},
  year={2018}
}

@article{resnet,
  title={Deep residual learning for image recognition. arXiv 2015},
  author={He, Kaiming and Zhang, Xiangyu and Ren, Shaoqing and Sun, Jian},
  journal={arXiv preprint arXiv:1512.03385},
  year={2015}
}

@inproceedings{cil,
  title={Class-incremental learning via deep model consolidation},
  author={Zhang, Junting and Zhang, Jie and Ghosh, Shalini and Li, Dawei and Tasci, Serafettin and Heck, Larry and Zhang, Heming and Kuo, C-C Jay},
  booktitle={Proceedings of the IEEE/CVF Winter Conference on Applications of Computer Vision},
  pages={1131--1140},
  year={2020}
}

@inproceedings{Exstream,
  title={Memory efficient experience replay for streaming learning},
  author={Hayes, Tyler L and Cahill, Nathan D and Kanan, Christopher},
  booktitle={2019 International Conference on Robotics and Automation (ICRA)},
  pages={9769--9776},
  year={2019},
  organization={IEEE}
}

@article{ExAI,
  title={Saliency Guided Experience Packing for Replay in Continual Learning},
  author={Saha, Gobinda and Roy, Kaushik},
  journal={arXiv preprint arXiv:2109.04954},
  year={2021}
}

@article{cil2,
  title={Class-incremental learning: survey and performance evaluation on image classification},
  author={Masana, Marc and Liu, Xialei and Twardowski, Bartlomiej and Menta, Mikel and Bagdanov, Andrew D and van de Weijer, Joost},
  journal={arXiv preprint arXiv:2010.15277},
  year={2020}
}

@article{mtl,
  title={A survey on multi-task learning},
  author={Zhang, Yu and Yang, Qiang},
  journal={IEEE Transactions on Knowledge and Data Engineering},
  year={2021},
  publisher={IEEE}
}

@inproceedings{yolo,
  title={You only look once: Unified, real-time object detection},
  author={Redmon, Joseph and Divvala, Santosh and Girshick, Ross and Farhadi, Ali},
  booktitle={Proceedings of the IEEE conference on computer vision and pattern recognition},
  pages={779--788},
  year={2016}
}

@article{gpt3,
  title={Language models are few-shot learners},
  author={Brown, Tom and Mann, Benjamin and Ryder, Nick and Subbiah, Melanie and Kaplan, Jared D and Dhariwal, Prafulla and Neelakantan, Arvind and Shyam, Pranav and Sastry, Girish and Askell, Amanda and others},
  journal={Advances in neural information processing systems},
  volume={33},
  pages={1877--1901},
  year={2020}
}

@article{CL_robotics,
title = {Continual learning for robotics: Definition, framework, learning strategies, opportunities and challenges},
journal = {Information Fusion},
volume = {58},
pages = {52-68},
year = {2020},
author = {Timothée Lesort and Vincenzo Lomonaco and Andrei Stoian and Davide Maltoni and David Filliat and Natalia Díaz-Rodríguez}
}

@article{LrL,
  title={Lifelong robot learning},
  author={Thrun, Sebastian and Mitchell, Tom M},
  journal={Robotics and autonomous systems},
  volume={15},
  number={1-2},
  pages={25--46},
  year={1995},
  publisher={Elsevier}
}

@article{CL-RL,
  title={Continual learning in reinforcement environments},
  author={Ring, Mark Bishop and others},
  year={1994},
  publisher={Citeseer}
}

@article{icarll,  author={Rebuffi, Sylvestre-Alvise and Kolesnikov, Alexander and Sperl, Georg and Lampert, Christoph H.},  journal={2017 IEEE Conference on Computer Vision and Pattern Recognition (CVPR)},   title={iCaRL: Incremental Classifier and Representation Learning},   year={2017}, pages={5533-5542}}

@article{RR,
  title={Catastrophic forgetting, rehearsal and pseudorehearsal},
  author={Robins, Anthony},
  journal={Connection Science},
  volume={7},
  number={2},
  pages={123--146},
  year={1995},
  publisher={Taylor \& Francis}
}

@inproceedings{avalanche,
  title={Avalanche: an end-to-end library for continual learning},
  author={Lomonaco, Vincenzo and Pellegrini, Lorenzo and Cossu, Andrea and Carta, Antonio and Graffieti, Gabriele and Hayes, Tyler L and De Lange, Matthias and Masana, Marc and Pomponi, Jary and Van de Ven, Gido M and others},
  booktitle={Proceedings of the IEEE/CVF Conference on Computer Vision and Pattern Recognition},
  pages={3600--3610},
  year={2021}
}

@article{DD,
  title={Dataset Distillation},
  author={Wang, Tongzhou and Zhu, Jun-Yan and Torralba, Antonio and Efros, Alexei A},
  journal={arXiv preprint arXiv:1811.10959},
  year={2018}
}

@article{
DCGM,
title={Dataset Condensation with Gradient Matching},
author={Bo Zhao and Konda Reddy Mopuri and Hakan Bilen},
journal={International Conference on Learning Representations},
year={2021},
url={https://openreview.net/forum?id=mSAKhLYLSsl}
}

@article{GEM,
  title={Gradient Episodic Memory for Continual Learning},
  author={David Lopez-Paz and Marc'Aurelio Ranzato},
  journal={NIPS},
  year={2017}
}

@article{A-GEM,
title={Efficient Lifelong Learning with A-{GEM}},
author={Arslan Chaudhry and Marc’Aurelio Ranzato and Marcus Rohrbach and Mohamed Elhoseiny},
journal={International Conference on Learning Representations},
year={2019},
url={https://openreview.net/forum?id=Hkf2_sC5FX},
}

@article{Gdumb,
author="Prabhu, Ameya
and Torr, Philip H. S.
and Dokania, Puneet K.",
editor="Vedaldi, Andrea
and Bischof, Horst
and Brox, Thomas
and Frahm, Jan-Michael",
title="GDumb: A Simple Approach that Questions Our Progress in Continual Learning",
journal="Computer Vision -- ECCV 2020",
year="2020",
publisher="Springer International Publishing",
address="Cham",
pages="524--540",
isbn="978-3-030-58536-5"
}

@article{MIR,
archivePrefix = {arXiv},
arxivId = {1908.04742},
author = {Aljundi, Rahaf and Caccia, Lucas and Belilovsky, Eugene and Caccia, Massimo and Lin, Min and Charlin, Laurent and Tuytelaars, Tinne},
journal = {Advances in Neural Information Processing Systems},
eprint = {1908.04742},
issn = {10495258},
title = {{Online continual learning with maximally interfered retrieval}},
volume = {volume 32},
year = {2019}
}

@article {EWC,
	author = {Kirkpatrick, James and Pascanu, Razvan and Rabinowitz, Neil and Veness, Joel and Desjardins, Guillaume and Rusu, Andrei A. and Milan, Kieran and Quan, John and Ramalho, Tiago and Grabska-Barwinska, Agnieszka and Hassabis, Demis and Clopath, Claudia and Kumaran, Dharshan and Hadsell, Raia},
	title = {Overcoming catastrophic forgetting in neural networks},
	volume = {114},
	number = {13},
	pages = {3521--3526},
	year = {2017},
	doi = {10.1073/pnas.1611835114},
	publisher = {National Academy of Sciences},
	issn = {0027-8424},
	URL = {https://www.pnas.org/content/114/13/3521},
	eprint = {https://www.pnas.org/content/114/13/3521.full.pdf},
	journal = {Proceedings of the National Academy of Sciences}
}

@article{LwF,
  author={Li, Zhizhong and Hoiem, Derek},
  journal={IEEE Transactions on Pattern Analysis and Machine Intelligence}, 
  title={Learning without Forgetting}, 
  year={2018},
  volume={40},
  number={12},
  pages={2935-2947},
  doi={10.1109/TPAMI.2017.2773081}}

@article{SI,
  title = 	 {Continual Learning Through Synaptic Intelligence},
  author =       {Friedemann Zenke and Ben Poole and Surya Ganguli},
  journal = 	 {Proceedings of the 34th International Conference on Machine Learning},
  pages = 	 {3987--3995},
  year = 	 {2017},
  editor = 	 {Precup, Doina and Teh, Yee Whye},
  volume = 	 {70},
  series = 	 {Proceedings of Machine Learning Research},
  month = 	 {06--11 Aug},
  publisher =    {PMLR},
  url = 	 {https://proceedings.mlr.press/v70/zenke17a.html},
}

@article{BiC,
  author={Wu, Yue and Chen, Yinpeng and Wang, Lijuan and Ye, Yuancheng and Liu, Zicheng and Guo, Yandong and Fu, Yun},
  journal={2019 IEEE/CVF Conference on Computer Vision and Pattern Recognition (CVPR)}, 
  title={Large Scale Incremental Learning}, 
  year={2019},
  volume={},
  number={},
  pages={374-382},
  doi={10.1109/CVPR.2019.00046}}

@article{three-scen,
       author = {{van de Ven}, Gido M. and {Tolias}, Andreas S.},
        title = "{Three scenarios for continual learning}",
      journal = {arXiv preprint arXiv:1904.07734},
         year = 2019,
        month = apr,
 primaryClass = {cs.LG}
}

@article{PNN,
  author    = {Andrei A. Rusu and
               Neil C. Rabinowitz and
               Guillaume Desjardins and
               Hubert Soyer and
               James Kirkpatrick and
               Koray Kavukcuoglu and
               Razvan Pascanu and
               Raia Hadsell},
  title     = {Progressive Neural Networks},
  journal   = {CoRR},
  year      = {2016},
  url       = {http://arxiv.org/abs/1606.04671},
  eprinttype = {arXiv},
  eprint    = {1606.04671},
  bibsource = {dblp computer science bibliography, https://dblp.org}
}

@article{ExpertGate,
author = {R. Aljundi and P. Chakravarty and T. Tuytelaars},
journal = {2017 IEEE Conference on Computer Vision and Pattern Recognition (CVPR)},
title = {Expert Gate: Lifelong Learning with a Network of Experts},
year = {2017},
volume = {},
issn = {1063-6919},
pages = {7120-7129},
doi = {10.1109/CVPR.2017.753},
url = {https://doi.ieeecomputersociety.org/10.1109/CVPR.2017.753},
publisher = {IEEE Computer Society},
address = {Los Alamitos, CA, USA},
month = {jul}
}

@article{HAT,
  title = 	 {Overcoming Catastrophic Forgetting with Hard Attention to the Task},
  author =       {Serra, Joan and Suris, Didac and Miron, Marius and Karatzoglou, Alexandros},
  journal = 	 {Proceedings of the 35th International Conference on Machine Learning},
  pages = 	 {4548--4557},
  year = 	 {2018},
  editor = 	 {Dy, Jennifer and Krause, Andreas},
  volume = 	 {80},
  series = 	 {Proceedings of Machine Learning Research},
  month = 	 {10--15 Jul},
  publisher =    {PMLR},
  url = 	 {https://proceedings.mlr.press/v80/serra18a.html},
}

@article{PackNet,
  author={Mallya, Arun and Lazebnik, Svetlana},
  journal={2018 IEEE/CVF Conference on Computer Vision and Pattern Recognition}, 
  title={PackNet: Adding Multiple Tasks to a Single Network by Iterative Pruning}, 
  year={2018},
  volume={},
  number={},
  pages={7765-7773},
  doi={10.1109/CVPR.2018.00810}}

@article{CCM,
       author = {{Wiewel}, Felix and {Yang}, Bin},
        title = "{Condensed Composite Memory Continual Learning}",
      journal = {arXiv preprint arXiv:2102.09890},
         year = 2021,
        month = feb,
 primaryClass = {cs.LG},
       adsurl = {https://ui.adsabs.harvard.edu/abs/2021arXiv210209890W}
}

@article{cat-forg2,
  title={Catastrophic forgetting in connectionist networks},
  author={French, Robert M},
  journal={Trends in cognitive sciences},
  volume={3},
  number={4},
  year={1999},
  publisher={Elsevier}
}

@article{catastrophic-forgetting,
title = "Catastrophic Interference in Connectionist Networks: The Sequential Learning Problem",
author = "Michael McCloskey and Cohen, {Neal J.}",
year = "1989",
month = jan,
day = "1",
doi = "10.1016/S0079-7421(08)60536-8",
language = "English (US)",
volume = "24",
pages = "109--165",
journal = "Psychology of Learning and Motivation - Advances in Research and Theory",
issn = "0079-7421",
publisher = "Academic Press Inc.",
number = "C",
}

\end{document}